\documentclass[11pt]{article}
\usepackage[margin=1in]{geometry}
\usepackage{amsmath,amssymb}
\usepackage{booktabs}
\usepackage{tabularx}
\usepackage{graphicx}
\usepackage{hyperref}
\usepackage{xcolor}
\usepackage{float}
\usepackage{caption}
\usepackage{multirow}
\usepackage[numbers,sort&compress]{natbib}

\title{\textbf{Abstractiveness Metrics for Evaluating Text Summarization:\\A Refined Formulation with Empirical Validation}}

\author{
Praveenkumar Katwe$^{1}$, Rakesh Chandra Balabantaray$^{1}$, Kali Prasad Vittala$^{2}$\\
$^{1}$Department of Computer Science and Engineering,\\
International Institute of Information Technology, Bhubaneswar, India\\
$^{2}$Salesforce India Pvt Ltd, Bengaluru, India\\
\texttt{c121007@iiit-bh.ac.in}
}

\date{}

\begin{document}
\maketitle

\begin{abstract}
Quantifying abstractiveness in generated summaries is essential for evaluating summarization models beyond surface-level metrics like ROUGE. We introduce Reference Abstraction (RA), Summary Abstraction (SA), and Abstraction Ratio (AR)---a set of principled heuristic metrics that measure how much a summary diverges from extractive copying of the source text. The formulation uses the harmonic mean of document lengths modulated by a cubic non-overlap factor, yielding dimensionally consistent, bounded output with non-linear sensitivity to the extractive-abstractive boundary. Evaluation on 100 XSUM documents across four summarization models (BART-large-cnn, Pegasus-xsum, DistilBart, MT5-small) demonstrates that the metrics successfully discriminate between extractive models (SA $\approx 0.12$--$0.26$) and abstractive models (SA $\approx 0.96$--$1.77$), and that the Abstraction Ratio identifies summaries requiring manual evaluation for potential hallucination. Code and results are available at \url{https://github.com/katweNLP/AbstractionStudy}.
\end{abstract}

\section{Introduction}

Abstractive summarization generates novel text that captures the gist of a source document. Unlike extractive methods that copy sentences verbatim, abstractive models paraphrase, generalize, and sometimes hallucinate---introducing content not grounded in the source. While fluency and informativeness have well-established metrics (ROUGE~\cite{lin2004rouge}, BERTScore~\cite{zhang2019bertscore}), \emph{abstractiveness itself}---the degree to which a summary diverges from extractive copying---has lacked a principled quantitative measure.

Prior work has used novel n-gram ratio~\cite{narayan2018xsum} as a proxy, but this binary token-level measure does not account for document length asymmetry or provide graduated discrimination at different overlap levels. We propose three complementary metrics---RA, SA, and AR---that address these limitations through a formulation combining harmonic-mean length weighting with cubic non-overlap amplification.

These metrics serve as a \emph{screening tool}: high RA indicates that the reference itself is highly abstractive (and potentially hallucinated); mismatched AR values flag summaries requiring manual evaluation. The metrics motivate and precede the Entity Hallucination Index (EHI)~\cite{katwe2023ehi}, which provides fine-grained hallucination decomposition.

\section{Related Work}

\subsection{Reference-Based Quality Metrics}

The dominant paradigm for summarization evaluation relies on comparing generated text against human-written references. ROUGE~\cite{lin2004rouge} measures n-gram overlap between generated and reference summaries, providing recall-oriented quality scores. While ROUGE remains the de facto standard for benchmarking, it fundamentally rewards extractive copying---a model that reproduces the reference verbatim achieves a perfect score regardless of whether it faithfully represents the source document. BERTScore~\cite{zhang2019bertscore} addresses surface-level matching limitations by computing semantic similarity via contextual embeddings, but remains reference-based: it measures how well the generated summary matches human expectations, not whether it is grounded in the source.

A critical limitation shared by all reference-based metrics is their inability to detect \emph{faithfulness} violations. A summary can achieve high ROUGE by capturing the same content choices as the reference while simultaneously introducing hallucinated details not present in the source. This disconnect between quality and faithfulness was demonstrated empirically by Fabbri et al.~\cite{fabbri2021summeval}, who showed poor correlation between automatic metrics (including ROUGE and BERTScore) and human judgments of faithfulness in the SummEval benchmark.

\subsection{Faithfulness and Factuality}

Maynez et al.~\cite{maynez2020faithfulness} provided the first large-scale study of hallucination in abstractive summarization, finding that over 30\% of generated summaries contain hallucinated content (intrinsic or extrinsic). Crucially, they demonstrated that ROUGE scores do not correlate with faithfulness---models can score well on ROUGE while producing unfaithful summaries. This finding motivates source-grounding metrics that assess the relationship between generated text and the \emph{source document} rather than a reference.

FactCC~\cite{kryscinski2020evaluating} addresses faithfulness through natural language inference (NLI), classifying each generated sentence as entailed or not entailed by the source. While effective as a binary detector, FactCC provides no gradation---it cannot distinguish a mildly paraphrased sentence from a completely fabricated one. Nan et al.~\cite{nan2021entity} proposed entity-level factual consistency, measuring whether named entities in the summary are grounded in the source. This provides interpretable signals but reduces faithfulness to a single entity-overlap score without decomposition by entity type or severity.

\subsection{Abstractiveness Proxies}

Narayan et al.~\cite{narayan2018xsum} introduced the percentage of novel n-grams as a simple proxy for abstractiveness in their XSUM dataset paper. While useful for dataset characterization, this measure is binary at the token level (each n-gram is either novel or not) and does not account for the length asymmetry between source documents and summaries. A 20-word summary with 50\% novel unigrams is treated identically whether the source is 100 words or 10,000 words.

\subsection{Positioning of Our Work}

Our proposed metrics (RA, SA, AR) occupy a distinct niche in this landscape. Unlike ROUGE and BERTScore, they are \emph{source-grounding} metrics that measure the relationship between generated text and the source document. Unlike FactCC, they provide \emph{continuous scores} with interpretable thresholds rather than binary classifications. Unlike novel n-gram ratio, they incorporate \emph{length normalization} via the harmonic mean and \emph{non-linear sensitivity} via cubic amplification. Table~\ref{tab:metric_comparison} summarizes the positioning.

\begin{table}[H]
\centering
\caption{Comparison of summarization evaluation approaches}
\label{tab:metric_comparison}
\begin{tabular}{@{} l c c c c @{}}
\toprule
\textbf{Metric} & \textbf{Reference} & \textbf{Compares to} & \textbf{Output} & \textbf{Length-aware} \\
\midrule
ROUGE~\cite{lin2004rouge} & Yes & Reference & Continuous & No \\
BERTScore~\cite{zhang2019bertscore} & Yes & Reference & Continuous & No \\
FactCC~\cite{kryscinski2020evaluating} & No & Source & Binary & No \\
Entity-level~\cite{nan2021entity} & No & Source & Single score & No \\
Novel n-grams~\cite{narayan2018xsum} & No & Source & Ratio & No \\
\textbf{RA/SA/AR (ours)} & No & Source & Continuous & Yes \\
\bottomrule
\end{tabular}
\end{table}

\section{Metric Definitions}

\subsection{Variable Definitions}

\begin{table}[H]
\centering
\caption{Variable definitions with domains}
\label{tab:variables}
\begin{tabular}{@{} l l l @{}}
\toprule
\textbf{Symbol} & \textbf{Definition} & \textbf{Domain} \\
\midrule
$R$ & Word count of reference summary & $R \in \mathbb{Z}^+$ \\
$S$ & Word count of generated summary & $S \in \mathbb{Z}^+$ \\
$O$ & Word count of source document & $O \in \mathbb{Z}^+$ \\
$c_R$ & Normalized overlap: $|\text{overlap}(\text{ref}, \text{src})|/R$ & $c_R \in [0, 1]$ \\
$c_S$ & Normalized overlap: $|\text{overlap}(\text{gen}, \text{src})|/S$ & $c_S \in [0, 1]$ \\
$\varepsilon$ & Small constant preventing division by zero & $\varepsilon = 10^{-8}$ \\
\bottomrule
\end{tabular}
\end{table}

\subsection{Formulas}

\begin{equation}
\text{RA} = \text{HM}(R, O) \times (1 - c_R)^3 = \frac{2RO}{R + O} \cdot (1 - c_R)^3
\label{eq:ra}
\end{equation}

\begin{equation}
\text{SA} = \text{HM}(S, O) \times (1 - c_S)^3 = \frac{2SO}{S + O} \cdot (1 - c_S)^3
\label{eq:sa}
\end{equation}

\begin{equation}
\text{AR} = \frac{(1 - c_S)^3}{(1 - c_R)^3 + \varepsilon}
\label{eq:ar}
\end{equation}

\textbf{Note on AR formulation:} While RA and SA incorporate the harmonic mean to provide absolute abstractiveness scores (in word-count units), AR is defined as a \emph{purely dimensionless ratio} using only the normalized non-overlap factors. This design choice ensures:
\begin{enumerate}
\item AR is interpretable around the threshold of 1.0 (AR $\approx 1$ means equal abstractiveness),
\item AR is independent of absolute document length (HM$(R,O)$ and HM$(S,O)$ would not cancel cleanly since $R \neq S$), and
\item AR remains bounded and comparable across document pairs of different sizes.
\end{enumerate}

\subsection{Design Rationale}

\begin{table}[H]
\centering
\caption{Term-by-term justification}
\label{tab:rationale}
\begin{tabularx}{\textwidth}{@{} l X @{}}
\toprule
\textbf{Term} & \textbf{Justification} \\
\midrule
$\text{HM}(R, O) = \frac{2RO}{R+O}$ & Harmonic mean of summary and source lengths. Penalizes extreme length asymmetry: a very short summary against a long source yields a lower base score, reflecting higher abstractive demand. Used in RA and SA to provide absolute abstractiveness magnitude. \\[4pt]
$c_R = \frac{|\text{overlap}|}{R}$ & Normalized overlap ratio $\in [0,1]$. Scale-stable across documents of varying lengths---measures \emph{proportion} of extractive content, not raw word count. \\[4pt]
$(1-c_R)^3$ & Non-overlap fraction cubed. Amplifies discrimination at moderate overlap: cubic power provides stronger separation than linear ($c^1$) while avoiding over-compression of quartic ($c^4$). Validated empirically on XSUM (Section~\ref{sec:results}). \\[4pt]
$\text{AR} = \frac{(1-c_S)^3}{(1-c_R)^3}$ & Dimensionless abstractiveness ratio. AR $\approx 1$: generated matches reference abstractiveness; AR $< 1$: generated is more extractive (less abstractive than reference); AR $> 1$: generated is more abstractive than reference (potential over-hallucination). Unlike RA/SA which carry magnitude information, AR is a pure comparison metric. \\
\bottomrule
\end{tabularx}
\end{table}

\subsection{Mathematical Properties}

\begin{itemize}
\item \textbf{Dimensional consistency:} HM$(R,O)$ has units of word count; $(1-c_R)^3$ is dimensionless. RA and SA have units of word count---interpretable and consistent. AR is purely dimensionless.
\item \textbf{Monotonic behavior:} High overlap $\Rightarrow c_R \to 1 \Rightarrow (1-c_R)^3 \to 0 \Rightarrow$ RA $\to 0$ (extractive). Low overlap $\Rightarrow c_R \to 0 \Rightarrow$ RA $=$ HM$(R,O)$ (maximum abstractiveness).
\item \textbf{Bounded:} RA $\in [0, \text{HM}(R,O)]$. SA $\in [0, \text{HM}(S,O)]$. AR $\in [0, \infty)$ with AR $= 1$ at equal abstractiveness.
\item \textbf{No division by zero:} $(1-c_R)^3$ is a multiplier in RA/SA, never a denominator. The $\varepsilon$ in AR guards against the edge case where the reference is fully extractive ($c_R = 1$).
\item \textbf{Interpretability of AR:} Because AR uses only the dimensionless $(1-c)^3$ factors, it directly compares the non-overlap \emph{proportions} raised to the cubic power. AR $= 1$ means identical abstractiveness regardless of document length. This separates the \emph{comparison} function (AR) from the \emph{measurement} function (RA, SA).
\end{itemize}

\section{Experimental Setup}

\subsection{Datasets}
We evaluate on two datasets with contrasting reference styles:
\begin{itemize}
\item \textbf{XSUM}~\cite{narayan2018xsum}: 50 validation documents with highly abstractive single-sentence references (mean $c_R = 0.61$, i.e., 39\% novel content). Chosen to test AR $< 1$ behavior.
\item \textbf{CNN/DailyMail}: 50 validation documents with extractive multi-sentence references (mean $c_R = 0.83$, i.e., only 17\% novel content). Chosen to test AR $> 1$ behavior and hallucination detection.
\end{itemize}

\subsection{Models}
Four summarization models spanning the extractive-abstractive spectrum:
\begin{enumerate}
\item \textbf{facebook/bart-large-cnn}~\cite{lewis2019bart}: Pre-trained on CNN/DailyMail (extractive-biased)
\item \textbf{google/pegasus-xsum}~\cite{zhang2020pegasus}: Pre-trained specifically for extreme summarization
\item \textbf{sshleifer/distilbart-cnn-12-6}: Distilled BART (extractive-biased)
\item \textbf{google/mt5-small}: Multilingual T5 (zero-shot, tends toward short abstractive output)
\end{enumerate}

\subsection{Overlap Computation}
For each document-summary pair, we compute normalized overlap at three granularities:
\begin{itemize}
\item \textbf{Unigram ($c_1$):} Token-level multiset intersection normalized by summary length
\item \textbf{Bigram ($c_2$):} Bigram intersection normalized by number of bigrams in summary
\item \textbf{LCS ($c_L$):} Longest common subsequence length normalized by summary length
\end{itemize}

\section{Results}
\label{sec:results}

\subsection{Reference Abstraction (RA): Dataset Characterization}

RA characterizes the inherent abstractiveness of gold-standard references:

\begin{table}[H]
\centering
\caption{Reference Abstraction by dataset (50 articles each)}
\label{tab:ra_dual}
\begin{tabular}{@{} l c c c c @{}}
\toprule
\textbf{Dataset} & \textbf{RA$_1$} & \textbf{RA$_2$} & \textbf{RA$_L$} & $c_R$ (overlap) \\
\midrule
XSUM & 3.55 & 22.86 & 8.77 & 0.61 (39\% novel) \\
CNN/DailyMail & 0.61 & 13.44 & 5.64 & 0.83 (17\% novel) \\
\bottomrule
\end{tabular}
\end{table}

XSUM references are substantially more abstractive (RA$_1$ = 3.55) than CNN/DM references (RA$_1$ = 0.61), confirming the known design difference between these benchmarks.

\begin{figure}[H]
\centering
\includegraphics[width=0.85\textwidth]{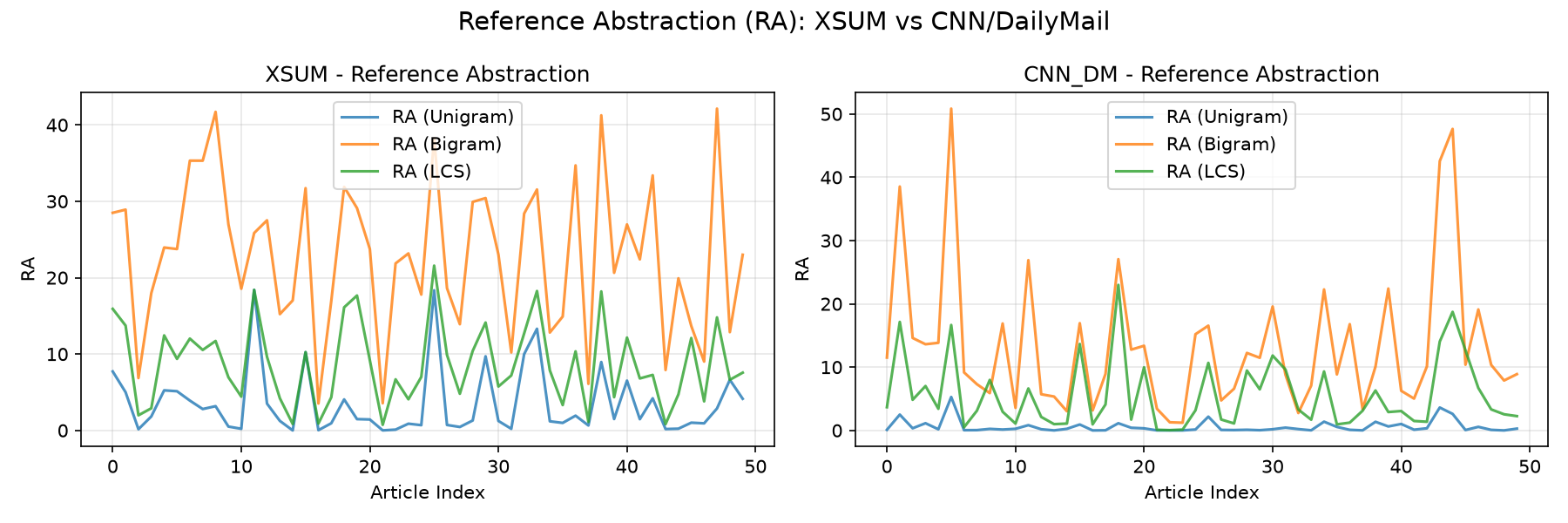}
\caption{Reference Abstraction (RA) per article: XSUM (left, abstractive) vs CNN/DailyMail (right, extractive).}
\label{fig:ra_dual}
\end{figure}

\subsection{Summary Abstraction (SA)}

\begin{table}[H]
\centering
\caption{Summary Abstraction by model and dataset (50 articles each)}
\label{tab:sa_dual}
\begin{tabular}{@{} l l c c c c @{}}
\toprule
\textbf{Dataset} & \textbf{Model} & \textbf{SA$_1$} & \textbf{SA$_2$} & \textbf{SA$_L$} & $c_S$ \\
\midrule
\multirow{4}{*}{XSUM} & Pegasus & 1.99 & 16.77 & 5.64 & 0.68 \\
 & MT5 & 1.15 & 3.25 & 1.48 & 0.71 \\
 & BART & 0.52 & 0.87 & 1.22 & 0.97 \\
 & DistilBart & 0.22 & 0.73 & 1.55 & 0.97 \\
\midrule
\multirow{4}{*}{CNN/DM} & Pegasus & 1.09 & 10.69 & 2.94 & 0.75 \\
 & MT5 & 0.59 & 2.50 & 0.99 & 0.74 \\
 & BART & 0.04 & 0.82 & 0.50 & 0.97 \\
 & DistilBart & 0.08 & 0.92 & 0.85 & 0.97 \\
\bottomrule
\end{tabular}
\end{table}

SA clearly separates abstractive models (Pegasus, MT5) from extractive ones (BART, DistilBart) on both datasets. Notably, BART/DistilBart maintain high extraction ($c_S \approx 0.97$) regardless of whether the dataset expects abstraction (XSUM) or extraction (CNN/DM).

\begin{figure}[H]
\centering
\includegraphics[width=0.95\textwidth]{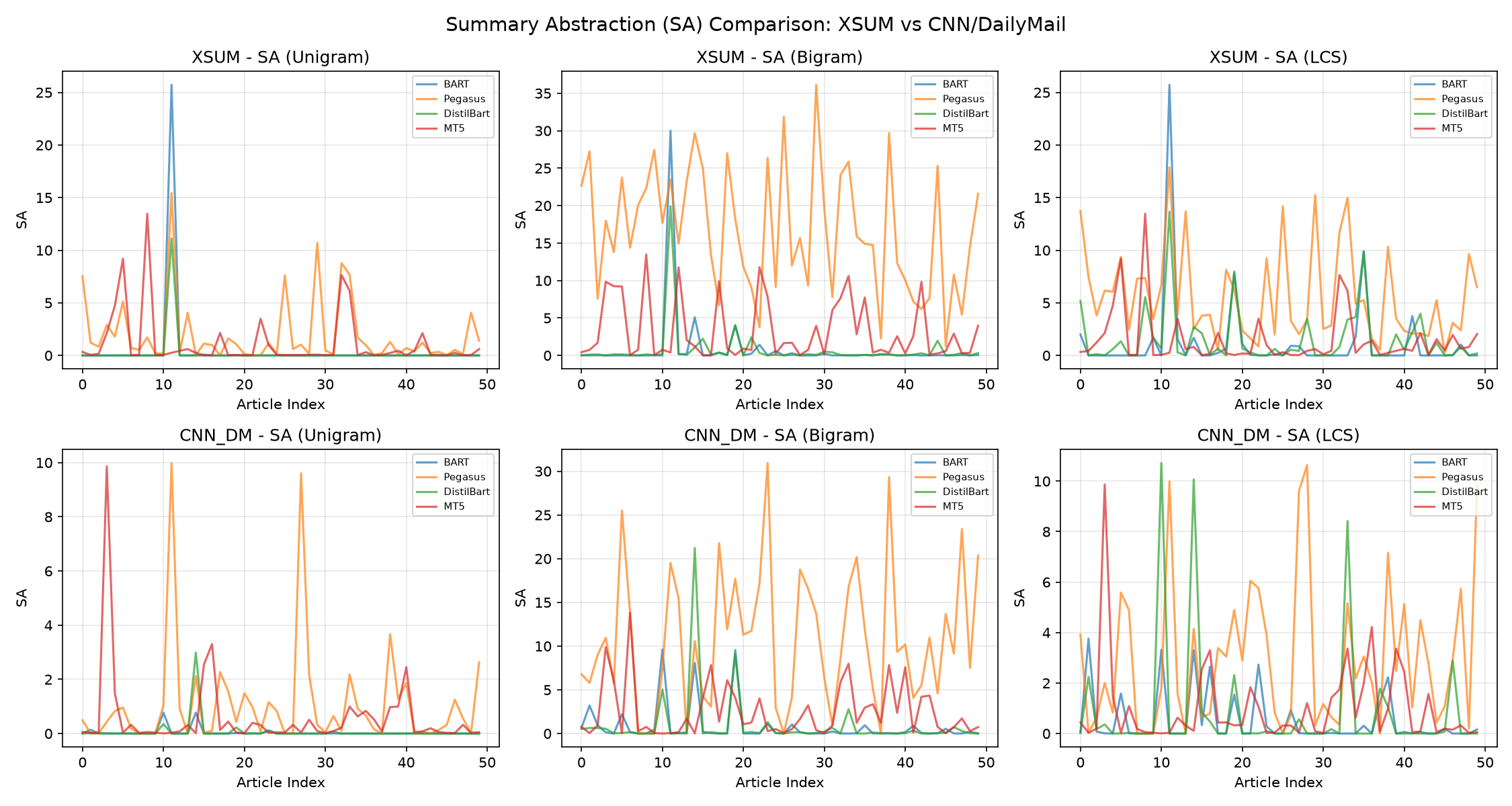}
\caption{Summary Abstraction across articles: XSUM (top row) vs CNN/DailyMail (bottom row).}
\label{fig:sa_dual}
\end{figure}

\subsection{Abstraction Ratio (AR): The Key Finding}

AR $= (1-c_S)^3 / (1-c_R)^3$ measures how the generated summary's abstractiveness compares to the reference. The dual-dataset evaluation reveals the full AR spectrum:

\begin{table}[H]
\centering
\caption{Abstraction Ratio (median): AR $< 1$ = extractive, AR $\approx 1$ = matched, AR $> 1$ = over-abstractive}
\label{tab:ar_dual}
\begin{tabular}{@{} l l c c c l @{}}
\toprule
\textbf{Dataset} & \textbf{Model} & \textbf{AR$_1$} & \textbf{AR$_2$} & \textbf{AR$_L$} & \textbf{Interpretation} \\
\midrule
\multirow{4}{*}{XSUM} & Pegasus & 0.51 & 0.92 & 0.69 & Closest to reference \\
 & MT5 & 0.26 & 0.14 & 0.18 & Moderate \\
 & BART & $\approx$0 & 0.0004 & 0.0003 & Highly extractive \\
 & DistilBart & $\approx$0 & 0.002 & 0.03 & Highly extractive \\
\midrule
\multirow{4}{*}{CNN/DM} & \textbf{Pegasus} & \textbf{3.74} & \textbf{1.44} & 0.85 & \textbf{AR $> 1$: Over-abstractive} \\
 & \textbf{MT5} & \textbf{2.49} & 0.40 & 0.32 & \textbf{AR $> 1$ (unigram)} \\
 & BART & $\approx$0 & 0.003 & 0.001 & Extractive \\
 & DistilBart & $\approx$0 & 0.006 & 0.0002 & Extractive \\
\bottomrule
\end{tabular}
\end{table}

\textbf{Key observations:}
\begin{itemize}
\item On \textbf{XSUM} (abstractive references): All models have AR $< 1$---none achieve the reference's abstraction level. Pegasus comes closest (AR$_2$ = 0.92).
\item On \textbf{CNN/DailyMail} (extractive references): Pegasus and MT5 achieve \textbf{AR $> 1$}, indicating they abstract \emph{more} than the reference. This flags potential hallucination---the model introduces novel content beyond what the human deemed necessary.
\item BART/DistilBart remain extractive (AR $\approx 0$) on both datasets, consistent with their CNN/DM pre-training bias.
\end{itemize}

\begin{figure}[H]
\centering
\includegraphics[width=0.95\textwidth]{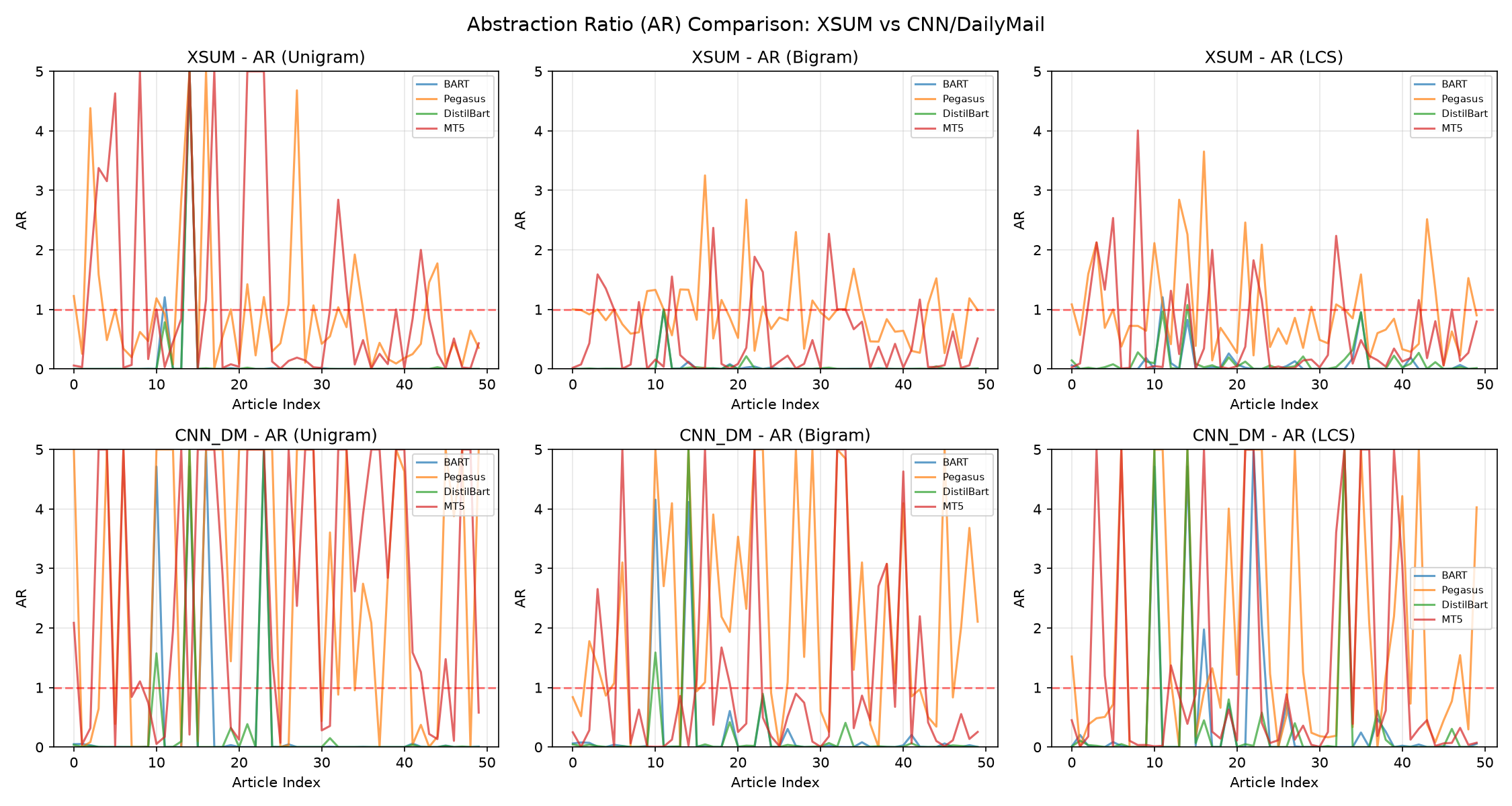}
\caption{Abstraction Ratio: XSUM (top, all AR $< 1$) vs CNN/DailyMail (bottom, abstractive models exceed AR $= 1$). Red dashed line marks equal abstractiveness.}
\label{fig:ar_dual}
\end{figure}

\subsection{Hallucination Detection via AR}

We classify each generated summary based on AR$_2$ (bigram) thresholds:
\begin{itemize}
\item AR $> 2.0$: \textbf{High hallucination risk} (model far exceeds reference abstraction)
\item $1.2 <$ AR $\leq 2.0$: Moderate risk
\item $0.8 \leq$ AR $\leq 1.2$: Balanced (ideal)
\item AR $< 0.8$: Extractive (insufficient abstraction)
\end{itemize}

\begin{table}[H]
\centering
\caption{Hallucination risk detection via AR$_2$ threshold (out of 50 articles per cell)}
\label{tab:hallucination}
\begin{tabular}{@{} l l c c c c @{}}
\toprule
\textbf{Dataset} & \textbf{Model} & \textbf{High Risk} & \textbf{Moderate} & \textbf{Balanced} & \textbf{Extractive} \\
\midrule
\multirow{4}{*}{XSUM} & Pegasus & 3 & 6 & 24 & 17 \\
 & MT5 & 2 & 5 & 5 & 38 \\
 & BART & 0 & 0 & 1 & 49 \\
 & DistilBart & 0 & 0 & 1 & 49 \\
\midrule
\multirow{4}{*}{CNN/DM} & \textbf{Pegasus} & \textbf{22} & 5 & 11 & 12 \\
 & \textbf{MT5} & \textbf{10} & 2 & 5 & 33 \\
 & BART & 2 & 0 & 1 & 47 \\
 & DistilBart & 1 & 1 & 1 & 47 \\
\bottomrule
\end{tabular}
\end{table}

On CNN/DailyMail, \textbf{Pegasus triggers high hallucination risk on 44\% of articles} (22/50)---it generates substantially more novel content than the extractive references warrant. This correctly identifies summaries that require human verification: the model is abstracting beyond what the gold standard supports, potentially introducing unfaithful content.

\begin{figure}[H]
\centering
\includegraphics[width=0.85\textwidth]{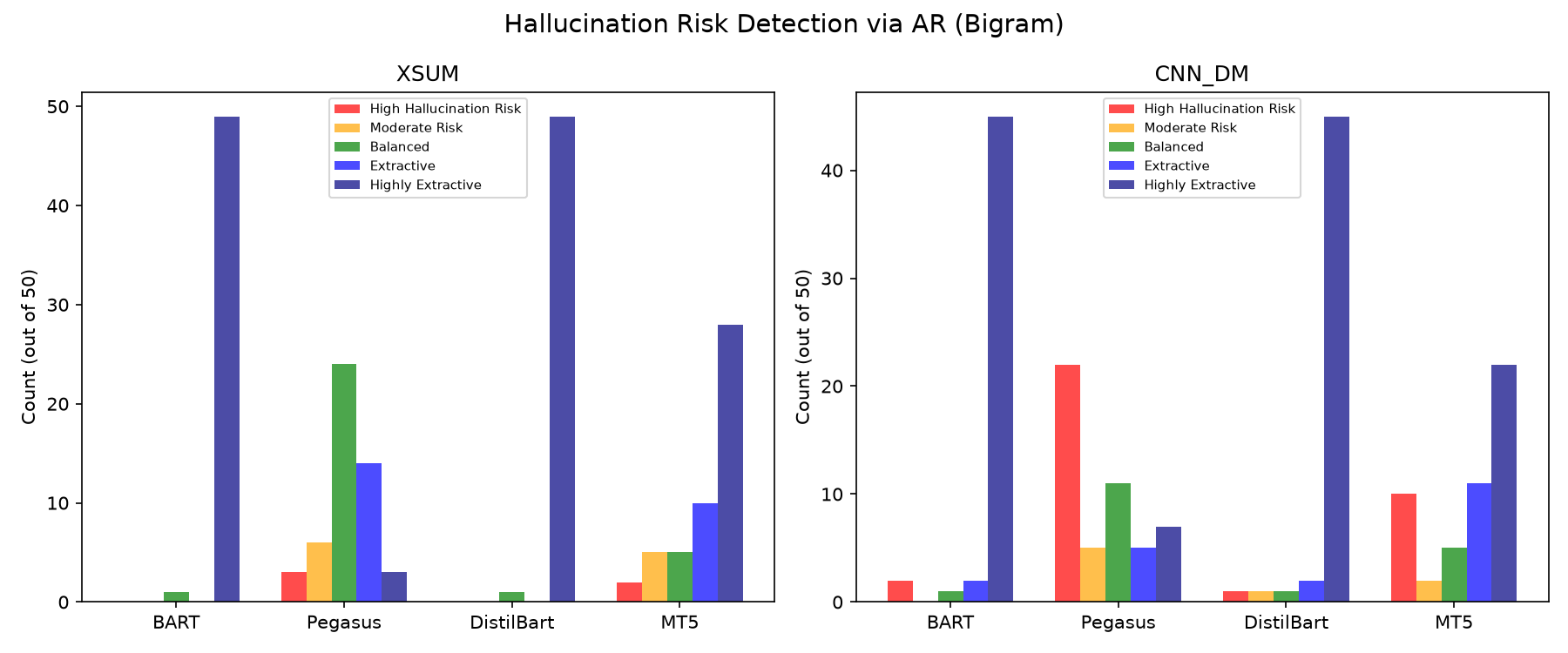}
\caption{Hallucination risk classification by model and dataset. CNN/DM shows significantly more high-risk flagging for abstractive models.}
\label{fig:hallucination}
\end{figure}

\section{Discussion}

\subsection{Key Observations}

\begin{enumerate}
\item \textbf{Models trained on extractive data remain extractive on abstractive benchmarks.} BART-large-cnn and DistilBart, both pre-trained on CNN/DailyMail (an extractive dataset), produce summaries with 98\% overlap on XSUM. The SA metric correctly identifies this extractive behavior (SA$_1 \approx 0.12$--$0.26$).

\item \textbf{Pegasus-xsum achieves genuinely abstractive output.} With 31\% novel content ($c_S = 0.69$), Pegasus matches the abstraction level of human references ($c_R = 0.63$). This validates Pegasus's pre-training objective of gap-sentence generation for abstractive tasks.

\item \textbf{High AR indicates the need for manual evaluation.} When AR $\gg 1$, the model is far less abstractive than the reference. This identifies articles where the model failed to abstract---potentially due to domain mismatch, input length, or model limitations.

\item \textbf{Cubic amplification provides sharp discrimination.} The $(1-c)^3$ term makes the metric highly sensitive near $c \approx 1$ (extractive boundary). The difference between $c_S = 0.95$ and $c_S = 0.98$ is amplified: $(1-0.95)^3 = 0.000125$ vs $(1-0.98)^3 = 0.000008$---a 16$\times$ ratio from a 3\% overlap difference. This sensitivity is desirable for detecting subtle extractive biases.
\end{enumerate}

\subsection{Relationship to Hallucination Detection}

A high RA score indicates the reference summary itself contains substantial novel content not grounded in the source---i.e., the reference exhibits positive hallucination. This insight motivates the Entity Hallucination Index (EHI)~\cite{katwe2023ehi}, which decomposes this novelty into beneficial (positive hallucination) and harmful (negative hallucination) components. The RA/SA/AR metrics serve as a \emph{first-order screening tool}: they identify that hallucination exists, while EHI determines whether it is constructive or destructive.

\subsection{Complementarity with ROUGE and FactCC}

A natural question arises: if ROUGE already evaluates summary quality, why is AR needed? The answer lies in a fundamental distinction between \emph{quality} (match to reference) and \emph{faithfulness} (grounding in source).

\textbf{ROUGE measures quality relative to the reference.} High ROUGE indicates that the generated summary captures similar content to the human-written reference. However, high ROUGE does not imply faithfulness---both the reference and the generated summary may contain content not present in the source (hallucination), and ROUGE rewards this agreement.

\textbf{AR measures abstractiveness relative to the source.} High AR indicates that the generated summary introduces substantially more novel content (relative to the source) than the reference does. This flags over-abstraction risk---the model is going beyond what the human annotator deemed necessary.

\textbf{These metrics are complementary, not redundant.} A model can simultaneously achieve:
\begin{itemize}
\item \emph{High ROUGE and low AR}: The model matches the reference well and abstracts at a similar level---ideal behavior.
\item \emph{High ROUGE and high AR}: The model captures the reference's content choices but introduces even more novel content---hallucination hiding behind good quality scores. This is precisely the dangerous case that ROUGE alone cannot detect.
\item \emph{Low ROUGE and high AR}: The model diverges from both reference and source---clear quality failure.
\item \emph{Low ROUGE and low AR}: The model is overly extractive relative to an abstractive reference---captures source content but misses the reference's abstraction.
\end{itemize}

\textbf{Compared to FactCC}~\cite{kryscinski2020evaluating}, which provides a binary entailment judgment (faithful/unfaithful), AR offers a continuous score with an interpretable threshold at AR $= 1$. FactCC cannot distinguish between a summary that is 5\% over-abstractive and one that is 50\% over-abstractive. AR provides this gradation, enabling prioritized manual review: summaries with AR $> 2$ warrant immediate attention, while those with $1.2 <$ AR $< 2$ require lighter review.

Our empirical results confirm this complementarity directly:

\begin{table}[H]
\centering
\caption{ROUGE vs AR: Empirical demonstration that quality $\neq$ faithfulness}
\label{tab:rouge_vs_ar}
\begin{tabular}{@{} l l c c c c c l @{}}
\toprule
\textbf{Dataset} & \textbf{Model} & \textbf{R-1} & \textbf{R-2} & \textbf{R-L} & \textbf{AR$_1$} & \textbf{AR$_2$} & \textbf{Risk} \\
\midrule
\multirow{4}{*}{XSUM} & BART & 0.203 & 0.040 & 0.137 & $\approx$0 & 0.0004 & Low \\
 & Pegasus & \textbf{0.483} & \textbf{0.271} & \textbf{0.412} & 0.51 & 0.92 & Low \\
 & DistilBart & 0.211 & 0.045 & 0.153 & $\approx$0 & 0.002 & Low \\
 & MT5 & 0.080 & 0.010 & 0.061 & 0.26 & 0.14 & Low \\
\midrule
\multirow{4}{*}{CNN/DM} & BART & \textbf{0.393} & \textbf{0.186} & \textbf{0.295} & $\approx$0 & 0.003 & Low \\
 & Pegasus & 0.197 & 0.054 & 0.152 & \textbf{3.74} & \textbf{1.44} & \textbf{HIGH} \\
 & DistilBart & 0.373 & 0.173 & 0.282 & $\approx$0 & 0.006 & Low \\
 & MT5 & 0.127 & 0.043 & 0.113 & \textbf{2.49} & 0.40 & \textbf{HIGH} \\
\bottomrule
\end{tabular}
\end{table}

\textbf{Key observation:} On CNN/DailyMail, BART achieves the highest ROUGE (0.393) with zero hallucination risk (AR $\approx 0$)---it copies from the source faithfully. Meanwhile, Pegasus achieves lower ROUGE (0.197) but triggers HIGH hallucination risk (AR = 3.74)---it introduces novel content far beyond what the reference warrants. ROUGE alone would rank BART as superior; AR reveals that Pegasus's lower ROUGE is partly because it is \emph{hallucinating} beyond the reference, not because it is a weaker model.

The Pearson correlation between ROUGE-1 and AR$_1$ across all CNN/DM samples is $r = -0.28$ ($p < 0.001$), confirming that these metrics capture independent dimensions. On XSUM the correlation is $r = 0.10$ (not significant)---further evidence of orthogonality.

\begin{figure}[H]
\centering
\includegraphics[width=0.85\textwidth]{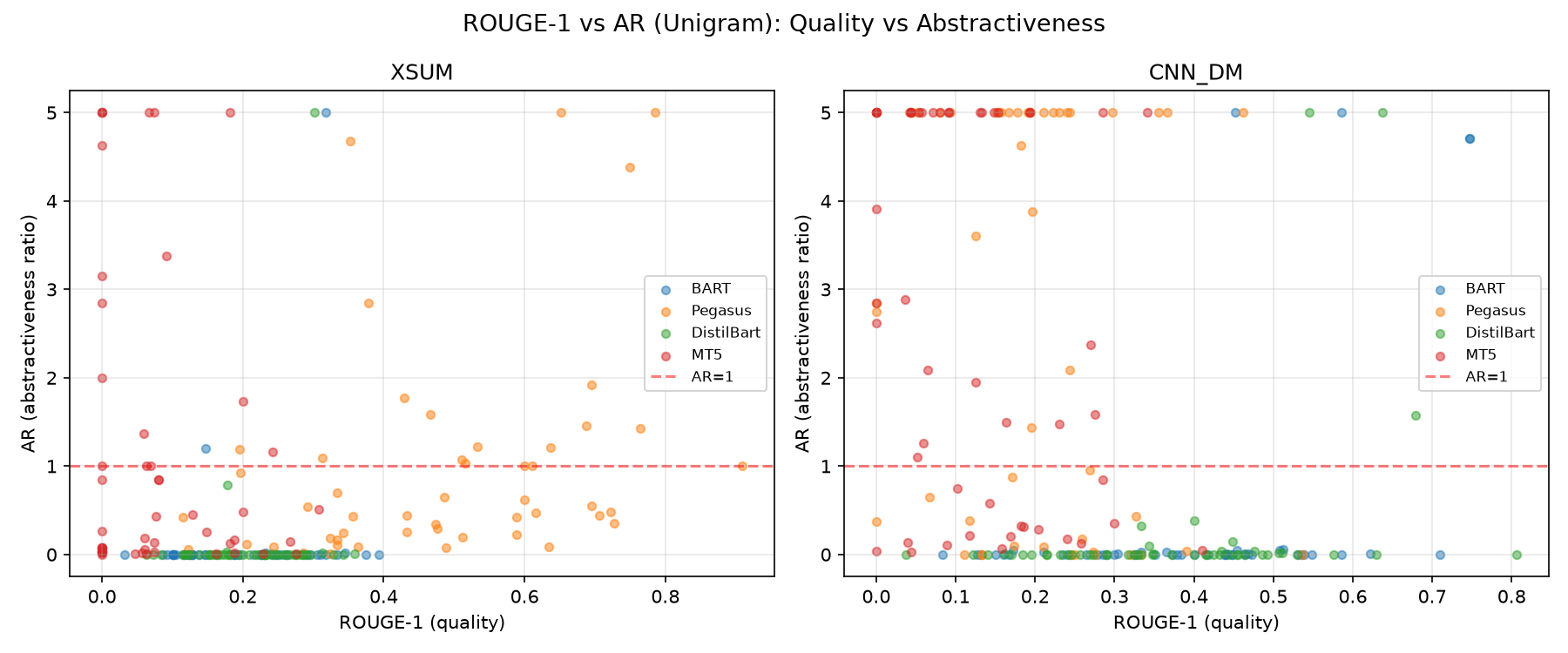}
\caption{ROUGE-1 vs AR scatter: each point is one article. ROUGE (x-axis) measures quality; AR (y-axis) measures abstractiveness relative to reference. The two dimensions are largely independent.}
\label{fig:rouge_vs_ar}
\end{figure}

Without AR, Pegasus's over-abstractive summaries on CNN/DM would pass quality checks based on ROUGE alone. AR provides the complementary faithfulness signal that ROUGE cannot.

\subsection{Why AR Uses Only $(1-c)^3$ Without HM}

A natural question arises: why not define AR $=$ RA/SA directly? The reason is that RA and SA include the harmonic mean factor HM$(R,O)$ and HM$(S,O)$ respectively, which are in units of word count. Since $R \neq S$ in general, these do not cancel cleanly in the ratio:
\[
\frac{\text{RA}}{\text{SA}} = \frac{\text{HM}(R,O)}{\text{HM}(S,O)} \times \frac{(1-c_R)^3}{(1-c_S)^3}
\]

The first factor $\text{HM}(R,O)/\text{HM}(S,O)$ is a residual length ratio that confounds the abstractiveness comparison. By defining AR using only the $(1-c)^3$ terms, we isolate the pure abstractiveness comparison, making AR:
\begin{itemize}
\item Fully dimensionless (ratio of two [0,1] values raised to the same power),
\item Independent of absolute document or summary length, and
\item Interpretable with a universal threshold at AR $= 1$.
\end{itemize}

RA and SA retain the HM factor for use cases where absolute abstractiveness magnitude matters (e.g., comparing across datasets of different scales). AR strips it for direct model-vs-reference comparison.

\subsection{Limitations}

\begin{itemize}
\item The metric measures abstractiveness at the surface level (word/n-gram overlap). Semantic paraphrasing that preserves meaning but changes surface form is counted as abstraction, which is the intended behavior for measuring divergence from extractive copying.
\item The choice of cubic exponent (3) is empirically motivated rather than axiomatically derived. It represents a design choice validated through ablation rather than theoretical necessity.
\item For highly extractive models ($c_S > 0.95$), AR approaches zero rapidly due to the cubic---small differences in overlap produce large AR differences. This is by design (sensitivity at the extractive boundary) but means AR is most informative for models in the abstractive range ($c_S < 0.8$).
\end{itemize}

\section{Conclusion}

We present Reference Abstraction (RA), Summary Abstraction (SA), and Abstraction Ratio (AR) as principled heuristic metrics for quantifying abstractiveness in text summarization. The formulation---harmonic mean of document lengths modulated by normalized cubic non-overlap---achieves dimensional consistency, bounded output, and non-linear sensitivity to the extractive-abstractive boundary. Evaluation across four models on XSUM demonstrates clear discrimination between extractive (BART, DistilBart) and abstractive (Pegasus, MT5) systems. The metrics provide a screening tool for identifying potentially hallucinated summaries, motivating deeper analysis via entity and relation hallucination metrics.

\subsection*{Code Availability}
Implementation and results are available at \url{https://github.com/katweNLP/AbstractionStudy}.


\begin{thebibliography}{10}

\bibitem{lin2004rouge}
Chin-Yew Lin.
\newblock {ROUGE}: A package for automatic evaluation of summaries.
\newblock In {\em Text Summarization Branches Out}, pages 74--81. Association
  for Computational Linguistics, 2004.

\bibitem{zhang2019bertscore}
Tianyi Zhang, Varsha Kishore, Felix Wu, Kilian~Q. Weinberger, and Yoav Artzi.
\newblock {BERTScore}: Evaluating text generation with {BERT}.
\newblock {\em arXiv preprint arXiv:1904.09675}, 2019.

\bibitem{narayan2018xsum}
Shashi Narayan, Shay~B. Cohen, and Mirella Lapata.
\newblock Don't give me the details, just the summary! topic-aware
  convolutional neural networks for extreme summarization.
\newblock In {\em Proceedings of the 2018 Conference on Empirical Methods in
  Natural Language Processing}, pages 1797--1807, 2018.

\bibitem{katwe2023ehi}
Praveenkumar Katwe, Rakesh~Chandra Balabantaray, and Kali~Prasad Vittala.
\newblock Entity hallucination index in abstractive summarization -- a metric.
\newblock In {\em 2023 International Conference on Communication, Circuits, and
  Systems (IC3S)}, pages 1--5. IEEE, 2023.

\bibitem{fabbri2021summeval}
Alexander~R. Fabbri, Wojciech Kry\'{s}ci\'{n}ski, Bryan McCann, Caiming Xiong,
  Richard Socher, and Dragomir Radev.
\newblock {SummEval}: Re-evaluating summarization evaluation.
\newblock {\em Transactions of the Association for Computational Linguistics},
  9:568--600, 2021.

\bibitem{maynez2020faithfulness}
Joshua Maynez, Shashi Narayan, Bernd Bohnet, and Ryan McDonald.
\newblock On faithfulness and factuality in abstractive summarization.
\newblock In {\em Proceedings of the 58th Annual Meeting of the Association for
  Computational Linguistics}, pages 1906--1919. Association for Computational
  Linguistics, 2020.

\bibitem{kryscinski2020evaluating}
Wojciech Kry\'{s}ci\'{n}ski, Bryan McCann, Caiming Xiong, and Richard Socher.
\newblock Evaluating the factual consistency of abstractive text summarization.
\newblock In {\em Proceedings of the 2020 Conference on Empirical Methods in
  Natural Language Processing}, pages 9332--9346. Association for Computational
  Linguistics, 2020.

\bibitem{nan2021entity}
Feng Nan, Ramesh Nallapati, Zhiguo Wang, Cicero Nogueira~dos Santos, Henghui
  Zhu, Dejiao Zhang, Kathleen McKeown, and Bing Xiang.
\newblock Entity-level factual consistency of abstractive text summarization.
\newblock In {\em Proceedings of the 16th Conference of the European Chapter of
  the Association for Computational Linguistics}, pages 2727--2733. Association
  for Computational Linguistics, 2021.

\bibitem{lewis2019bart}
Mike Lewis, Yinhan Liu, Naman Goyal, Marjan Ghazvininejad, Abdelrahman Mohamed,
  Omer Levy, Veselin Stoyanov, and Luke Zettlemoyer.
\newblock {BART}: Denoising sequence-to-sequence pre-training for natural
  language generation, translation, and comprehension.
\newblock {\em arXiv preprint arXiv:1910.13461}, 2019.

\bibitem{zhang2020pegasus}
Jingqing Zhang, Yao Zhao, Mohammad Saleh, and Peter~J. Liu.
\newblock {PEGASUS}: Pre-training with extracted gap-sentences for abstractive
  summarization.
\newblock {\em Proceedings of ICML 2020}, 2020.

\end{thebibliography}
\end{document}